\documentclass{article} 
\usepackage{iclr2026_delta,times}

\usepackage{amsmath,amsfonts,bm}

\def\eqref#1{equation~\ref{#1}}

\def\1{\bm{1}}

\DeclareMathAlphabet{\mathsfit}{\encodingdefault}{\sfdefault}{m}{sl}
\SetMathAlphabet{\mathsfit}{bold}{\encodingdefault}{\sfdefault}{bx}{n}

\usepackage{hyperref}
\usepackage{url}

\usepackage{amsmath}
\usepackage{amssymb}
\usepackage{mathtools}
\usepackage{amsthm}
\usepackage{algorithm}
\usepackage{algpseudocode}
\usepackage{booktabs}
\usepackage{caption}
\usepackage{multirow}
\usepackage{wrapfig}
\usepackage[capitalize,noabbrev]{cleveref}
\usepackage{tikz}
\usepackage[inkscapelatex=false]{svg}
\usetikzlibrary{positioning,arrows.meta,calc,matrix,fit,shapes.misc}

\theoremstyle{plain}
\newtheorem{theorem}{Theorem}[section]

\theoremstyle{definition}

\theoremstyle{remark}
\newtheorem{remark}[theorem]{Remark}

\newcommand{\cS}{\mathcal{S}}
\newcommand{\cT}{\mathcal{T}}
\newcommand{\cL}{\mathcal{L}}
\newcommand{\cU}{\mathcal{U}}
 
\def\d{\,\mathrm{d}}
\def\kl{\,\mathrm{KL}}

\iclrfinalcopy
\title{Self-Aware Markov Models\\ for Discrete Reasoning}

\author{Gregor Kornhardt\thanks{Equal contribution.} , Jannis Chemseddine\footnotemark[1] , Christian Wald \& Gabriele Steidl  \\
Technische Universität Berlin\\
Straße des 17. Juni 136, 10623 Berlin, Germany \\
\texttt{\{kornhardt, chemseddine, wald, steidl\}@math.tu-berlin.de} \\
}

\begin{document}

\maketitle

\begin{abstract}
Standard masked discrete diffusion models face limitations in reasoning tasks due to their inability to correct their own mistakes on the masking path. Since they rely on a fixed number of denoising steps, they are unable to adjust their computation to the complexity of a given problem. To address these limitations, we introduce a method based on learning a Markov transition kernel that is trained on its own outputs. This design enables tokens to be remasked, allowing the model to correct its previous mistakes. Furthermore, we do not need a fixed time schedule but use a trained stopping criterion. This allows for adaptation of the number of function evaluations to the difficulty of the reasoning problem. Our adaptation adds two lightweight prediction heads, enabling reuse and fine-tuning of existing pretrained models.
On the Sudoku-Extreme dataset we clearly outperform other flow based methods with a validity of $95\%$. For the Countdown-4 we only need in average of 10 steps to solve almost $96\%$ of them correctly, while many problems can be solved already in 2 steps.
\end{abstract}


\section{Introduction}

Chain-of-thought reasoning \cite{wei2023chainofthoughtpromptingelicitsreasoning} 
has become standard in autoregressive LLMs, but relies on sequential commitment, i.e. once a token is generated, possible errors propagate further. 
Discrete diffusion has been shown to be a scalable method \cite{nie2025largelanguagediffusionmodels, gat2024discrete, austin2023structureddenoisingdiffusionmodels, campbell2024generativeflowsdiscretestatespaces}, and offer offers a natural alternative, rather than committing sequentially, the model treats its output as a revised draft, where tokens can be sampled, reconsidered, and replaced. Although this flexibility could enable errors to be corrected rather than locked in, this potential remains largely unrealized so far.
In this paper, we are interested in reasoning tasks, where instead of generation problems, answers are uniquely determined, making error correction essential rather than optional. Typical examples are Sudoku and Countdown treated in our numerical part.
For these problems, current discrete flow matching training exhibits two key limitations, namely limited self-correction and non-adaptivity to the problem's complexity:
\begin{itemize}
\item[i)] \textbf{Self-Correction.} Under the standard masking learning process, a partially noised state contains either mask tokens or correct target tokens, but never incorrect ones. The model therefore learns to denoise only from idealized states. However, at inference, the model relies on states it has generated itself, which may contain errors. Once an incorrect token is placed, it persists, since standard samplers do not revisit unmasked positions. Consequently, subsequent predictions conditioned on a state, except for any seen during training, become unreliable.
\item[ii)] \textbf{Integration of Problem  Difficulty.} Standard inference uses a predetermined number of denoising steps. Therefore, instances that are easy for the model and could be solved quickly, still require the full budget, wasting computation and risking unnecessary errors from repeated resampling. On the other hand, hard instances that require more refinement receive no additional resources. In summary, without a mechanism to assess progress, the model cannot adapt its computation to the problem at hand.
\end{itemize}

\paragraph{Existing approaches.} Several strategies have been proposed to improve reasoning in discrete diffusion. Inference-time methods such as remasking \cite{wang2026remaskingdiscretediffusionmodels} and adaptive token ordering \cite{kim2025trainworstplanbest} modify the sampling procedure. The procedure enables error correction without retraining, but cannot change what the model has learned. Reinforcement learning methods \cite{zhao2025d1scalingreasoningdiffusion, tang2026wd1weightedpolicyoptimization} adapt policy gradient techniques to discrete diffusion, using reward signals to improve reasoning. This is a distinct paradigm from supervised training and introduces additional complexity such as reward design and training instability. Training-time loss modifications reweight tokens by difficulty \cite{ye2025autoregressiondiscretediffusioncomplex} or add auxiliary heads predicting per-token quality \cite{kim2025finetuningmaskeddiffusionprovable}. It improves learning, but does not  directly expose the model to states containing its own errors. GIDD of \cite{vonruette2025generalizedinterpolatingdiscretediffusion} generalizes the noise process to include uniform noise alongside masking, so that training states may contain random incorrect tokens. This exposes the model to \emph{some} errors, but not to its own systematic mistakes.
Finally, autoregressive models, also called next token-prediction are based on  
$p(x) = p(x_0)\prod_{i} p(x_i|x_{k<i})$
have many advantages in training, like KV-Caching \cite{pope2022efficientlyscalingtransformerinference}, and training on the whole sequence in parallel. However, they have the disadvantage that they can not generate in parallel and fix their own mistakes.

\begin{figure}[t]
    \centering
    \includegraphics[width=0.85\linewidth]{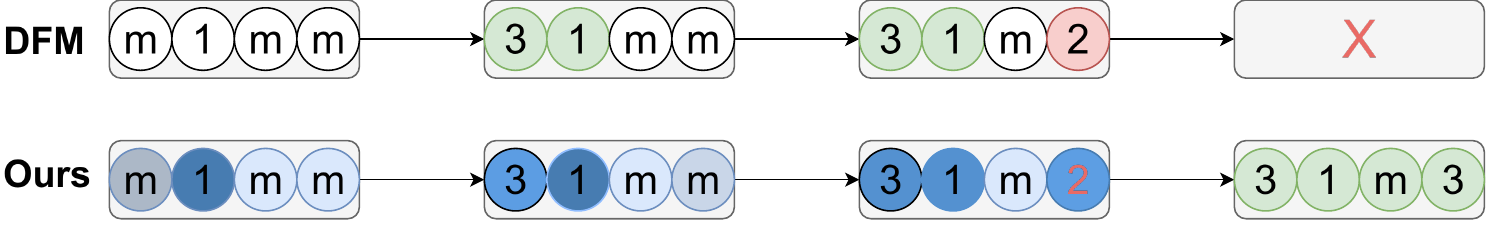}
    \caption{Inference trajectories. \textbf{DFM} \cite{gat2024discrete}: once the sampler leaves the masking path (red, incorrect token), it fails to recover and does not converge to the target solution (\(X\)). \textbf{Ours}: trained on model-induced off-path states, the model detects the mistake and corrects it, recovering the correct final sequence (green); color intensity indicates the model’s confidence.}
    \label{fig:idea}
\end{figure}

\paragraph{Our approach.} Motivated by discrete flow matching, we propose a training framework for reasoning tasks based on a simple principle: train the model on states it actually produces, with a mechanism to abstain when uncertain, in Figure \ref{fig:idea} this is visualized . This leads to two interdependent components. First, we learn a \emph{Markov transition kernel} that separates \emph{what} to predict from \emph{when} to commit: at each position, a learned confidence score gates whether the model outputs a concrete token or remains masked. Second, we train on \emph{on-policy states} generated by applying this mixed kernel, exposing the model to configurations containing its own errors. The training objective encourages both accuracy (predict the correct token) and safety (do not commit when uncertain). An auxiliary head estimates progress toward completion, enabling early stopping when the model is confident it has finished.

The resulting model is no longer a generative model in the probabilistic sense, since we do not maintain a valid probability path from noise to data, except when we have an unique pairing between clues and solution.  Our method is tailored to tasks where correctness matters more than distributional fidelity. In our model, discrete flow matching  inspires a iterative refinement mechanism and optimize directly for task performance.


\paragraph{Contributions.} Based on the identification of the above two limitations of standard discrete flow matching for reasoning tasks, namely that the model never trains on states containing its own errors, and that computation does not adapt to instance difficulty, we propose 

\begin{itemize}
    \item  a novel method combining a mixed prediction-commitment kernel with on-policy training, addressing both limitations within a supervised learning framework.
    First, we train on mistakes produced by the model, allowing it to self-correct. Second, we introduce a stopping time, allowing the model to spend more effort on harder problems. Finally, to solve easy problems faster, we introduce a confidence criterion that lets the model fill in easy tokens first, while handling incorrect tokens by re-masking or correcting them.
    
    \item We demonstrate improved reasoning performance on Sudoku-Extreme and Countdown compared to standard training and existing methods.
    \end{itemize}

\section{Self-Correcting, Difficulty-Aware Markov Model}
We want to model discrete data $\mathcal S = \mathcal T^d$, where $\mathcal T = \{T_1, \ldots, T_N\} \cup \{m\}$
is a set of tokens and $m$ is a special mask token.
States (i.e., samples) from $\mathcal S$ are denoted by
$x = (x^1, \ldots, x^d)$ with $x^i \in \mathcal T$.
By $\Delta_{\mathcal S}$, we denote the probability simplex with vertices $\mathcal S$.
In the following, we restrict ourselves to the reasoning setting, where there is a unique pairing between clues and solutions.
We denote the clue distribution by $X_0$, the solution distribution by $X_1$, and their pairing by $\pi$. Note that
$(x_0,x_1) \sim \pi$, if $x_1$ is the solution of $x_0$.
For example, in Sudoku, we have $\mathcal T = \{1, \ldots, 9\}\cup \{m\}$ and $d = 81$. The clues are the given numbers at a certain position, while the remaining fields are filled with mask tokens. 

\paragraph{Discrete-Time Markov Chain}
A homogeneos discrete-time Markov chain $(Y_n)_n$ of random variables $Y_n \in \mathcal S$  is characterized by its transition kernel \(P\) (transition matrix) with entries
$$
P(x|y):= \mathbb{P}(Y_{n}=x \mid Y_{n-1}=y).
$$
While most existing methods fix a forward (noising) kernel and learn a backward (denoising) kernel, see, e.g. \cite{austin2023structureddenoisingdiffusionmodels}, we instead learn the forward kernel directly. Our goal is that, for sufficiently large $n$ we have $P^{n} \delta_{x_0} \approx \delta_{x_1}$ for all $(x_0,x_1) \sim \pi$. Since we want to find a solution that matches the clue, we force our Markov kernel to leave the clues unchanged. Consequently our Markov kernel depends on $x_0$, which encodes the clues. For readability we omit this dependence from the notation and write $P$ instead of $P^{x_0}$.  

The Markov kernel should  be easy to model and learn, concentrate on the correct solution, and  enable self-correction and adaptive computation based on problem difficulty. To this end, we model a \textit{conditional target distribution} $p_1^\theta(\cdot\mid y)$, an \textit{adaptive confidence} $c^\theta(y)$ (see \ref{subsec: Confidence for Self-Correction}) that decides whether to keep a token masked, unmasked, or to re-mask it, and a \textit{stopping time} $\tau^\theta(y)$ (see \ref{subsec: Difficulty-Aware Computation Time}).
We propose to learn a network given by
\begin{equation}\label{eq:network}
T_\theta: \mathcal{S}\rightarrow \Delta_{\mathcal S} \times [0,1]^d \times [0,1],
\quad
y \mapsto (p_1^\theta(\cdot\mid y), c^\theta(y), \tau^\theta(y)),
\end{equation}
where we keep the standard assumption that we learn a product distribution of the form
$$
p_1^\theta(\cdot\mid y) \coloneqq \prod_{i=1}^d p_1^{\theta,i}(x^i\mid y).
$$

Based on our network, we define a Markov kernel $P^\theta$ by
\begin{align}\label{eq:markov_kernel}
P^\theta(x\mid y) \coloneqq
\left\{
\begin{array}{ll}
\delta_y(x) & \text{if } \tau^\theta(y) \ge 1-\varepsilon,\\[2mm]
\prod_{i=1}^d \Big((1-c^{\theta,i}(y))\,\delta_m(x^i) + c^{\theta,i}(y)\,p_1^{\theta,i}(x^i\mid y)\Big) & \text{otherwise}.
\end{array}
\right.
\end{align}
In practice we fix the clues, i.e. we force $P^{\theta}(x\mid y)=0$ if the clues of $x$ do not match the clues of $x_0$. 
If the confidence $c^{\theta,i}(y)$ is large, the probability of the token $x^i$ follows $p_1^{\theta,i}(x^i\mid y)$. If the confidence is small, the model may keep the position masked or re-mask it. We reapply the kernel until the stopping time is hit.

\paragraph{Inference.}
Let $\mu_0^\theta= P_{X_0}$ be the law of $X_0$. For inference, we start in $x_0\sim X_0$ admitting a solution, and iteratively update the chain as
\begin{align}\label{eq:inference}
x_{n+1} \sim P^\theta(\cdot\mid x_n), \quad
\mu^\theta_{n+1} = \mu^\theta_{n} P^\theta\coloneqq \sum_{y\in \mathcal S}P^\theta(\cdot\mid y)\mu_n^\theta(y).
\end{align}
Note that $\mu_n^\theta$ depends on $x_0$, since $P^{\theta}$ does as well.

\textbf{Training.}
We  train the network such that the stationary distribution of $P^\theta$ approximates $\delta_{x_1}$.

Training on the path distribution $(\mu_n^\theta)_n$ is expensive, since we can only obtain it by unrolling the network, i.e., performing a fixed number of  forward steps in~\eqref{eq:inference}. 
Instead we use ideas from discrete flow matching, see Appendix \ref{sec:DFM}.
We use the masking path\footnote{For a random variable $X_t \in \mathcal S$, $t \in [0,1]$ with probability $p_{X_t} (x) = \mathbb P(X_t = x)$, we just write $p_t(x)$, and accordingly $p_{t|0,1}(x|x_0,x_1) = p_{X_t|(X_0,X_1) = (x_0,x_1)}(x)$.}
\begin{align}\label{eq:masking path}
p_{t|0,1}^i(\cdot \mid x_0,x_1) := (1-\kappa_t)\,\delta_{m}^i(\cdot) + \kappa_t\,\delta_{x_1}^i(\cdot),
\end{align}
with a schedule function $\kappa_t: [0,1] \to [0,1]$, for example $\kappa_t = t$. We approximate $(\mu_n^\theta)_n$ by performing one Markov-kernel step on $p_{t|0,1}$, resulting in
$$\nu_t^\theta (\cdot\mid x_0,x_1) \coloneqq p_{t|0,1}(\cdot\mid x_0,x_1)\,P^\theta.$$

Allowing the network in ~\eqref{eq:network} to learn all components, we propose to minimize the following loss, where $\operatorname{sg}$ denotes stop grad:
\begin{equation}\label{eq:loss}
\begin{aligned}
\mathcal{L}(\theta) &= \mathbb{E}_{t\sim \mathcal{U}[0,1],\ (x_0,x_1)\sim \pi,\ z \sim \nu_t^{\operatorname{sg}(\theta)}(\cdot\mid x_0,x_1)} \Big[
-\sum_{i=1}^d \log \big(c^{i,\theta}(z)\, p_1^{\theta,i}(x_1^i\mid z)\big) 
\\
&\quad - \frac{1}{1-c^i_\theta(z)} \log \Big(1-c^{\theta,i}(z)\sum_{v\in \mathcal{T}\backslash\{m,x^i_1\}} p_1^{\theta,i}(v\mid z)\Big) 
\\
&\quad + \big|\tau(z|x_1,x_0) - \tau_\theta(z)\big|\bigg].
\end{aligned}
\end{equation}
In the following subsection, we explain our choice of the loss.
\subsection{Confidence for Self-Correction}\label{subsec: Confidence for Self-Correction}
Ideally, the model decides for itself whether to unmask a token, keep it unmasked, or ultimately remask it. To learn this behaviour, the model must encounter states $z^i$ that are neither correct nor masked, and learn the best strategy in such situations. One could introduce uniformly random mistakes into the interpolation path, as in GIDD~\cite{vonruette2025generalizedinterpolatingdiscretediffusion}, but this would waste training capacity on errors that the model would not have made anyway. Diffusion-Chain-of-Thought~\cite{ye2024diffusion} instead uses, for $5\%$ of the training states, a solution $x_1$ obtained by running the reverse process.

\begin{figure}
    \centering
    \includegraphics[width=0.80\linewidth]{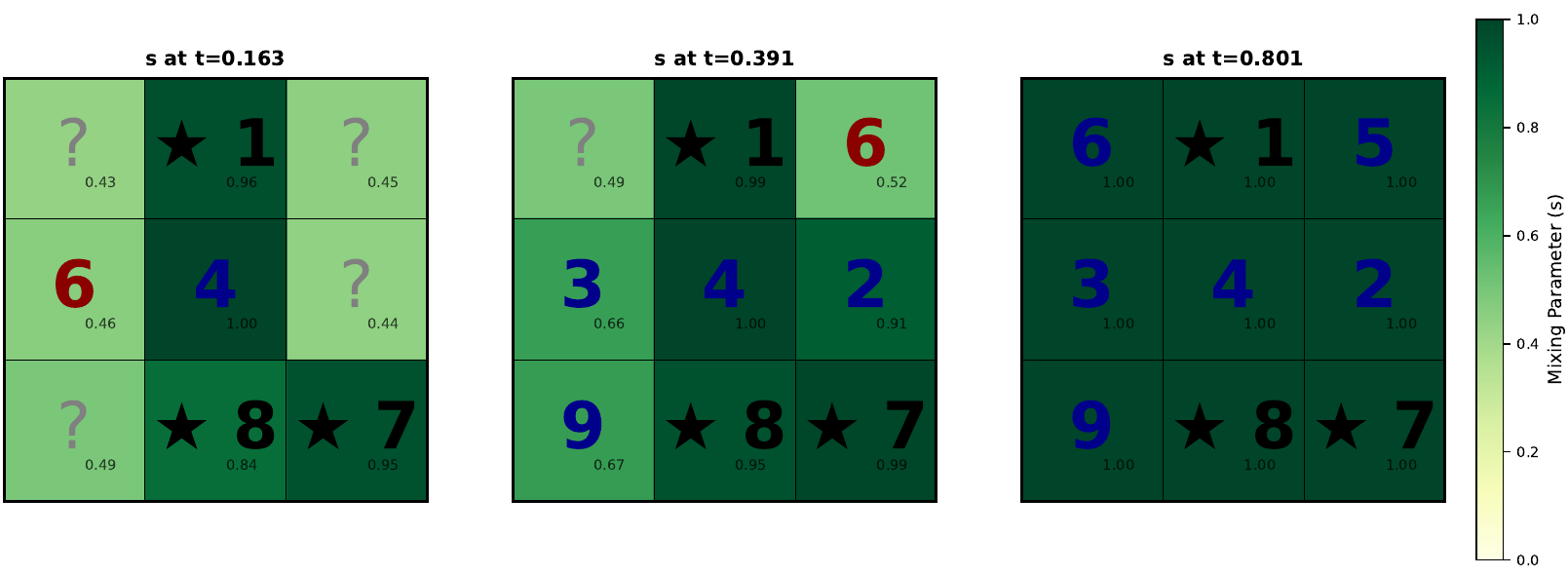}
\caption{Inference on a $3{\times}3$ Sudoku subgrid from a complete puzzle. We show the state after $k\!=\!1,2,3$ refinement steps; the predicted progress $\tau_\theta$ is displayed above each panel. Cell shading encodes confidence (mixing score $c_\theta$); clue cells are marked with $\ast$. Correct predictions are blue and incorrect ones red.}
    \label{fig:sudoku fix}
\end{figure}

Instead, we introduce model-dependent errors by taking one forward step with our Markov kernel, which yields $\nu_t^\theta$. On states $z \sim \nu_t^\theta$, the model can learn to estimate its own reliability based on its predictions $p^\theta_1(\cdot\mid z)$. Using our kernel~\eqref{eq:markov_kernel}, the model can progress faster when it is confident, this is visualized in \cref{fig:sudoku fix}. This contrasts with most diffusion models, which fix the number of inference steps. To obtain this behaviour, we maximize the probability of being correct via the term $-\sum_{i=1}^d \log\!\big(c^{i,\theta}(z)\, p_1^{\theta,i}(x_1^i\mid z)\big)$, while penalizing incorrect unmaskings $\frac{1}{1-c^i_\theta(z)} \log\!\left(1-c^{\theta,i}(z)\sum_{v\in \mathcal{T}\backslash\{m,x^i_1\}} p_1^{\theta,i}(v\mid z)\right)$ in ~\eqref{eq:loss}.

\subsection{Difficulty-Aware Computation Time}\label{subsec: Difficulty-Aware Computation Time}
Most discrete diffusion models use a fixed time schedule. The model must converge within a predetermined step budget. Instead, we want the model to spend more steps on harder instances.

We introduce a stopping time on an appropriate filtered probability space, which serves two purposes. First, it provides a valid stopping criterion. Second, it can be used to determine an adaptive step size.
To this end, we let the network also predict its own time, defined by
\begin{align}\label{eq:real time}
    \tau(x_t \mid x_1, x_0) =
    \frac{\big|\{x_t^i \mid \text{correct and not a clue}\}\big|}
         {\big|\{x_t^i \mid \text{not a clue}\}\big|}.
\end{align}

\begin{remark}
With regard to the flow-matching setting in Appendix~\ref{sec:DFM},
we see from~\eqref{eq:path} that
\begin{align}
\mathbb{E}_{x \sim p_{t\mid 0,1}}\big[\tau(x\mid x_0,x_1)\big]
&=
\sum_{x \,:\, x^i \in \{x_1^i, m\}} \tau(x\mid x_0, x_1)
\prod_{i=1}^d \Big((1-\kappa_t)\,\delta_m(x^i) + \kappa_t\,\delta_{x_1^i}(x^i)\Big)
= \kappa_t,
\end{align}
i.e., for $\kappa_t = t$, the expectation of $\tau(\cdot\mid x_0,x_1)$ equals the path time.
\end{remark}

\subsection{Inference Strategies}\label{subsec:Inference Strategies}
The method converges substantially faster (see Section~\ref{sec: Experiments}).
As a result, we can afford to run multiple sampling instances in parallel and aggregate them into a single prediction:
for each position, we take an $\arg\max$ over the empirical token frequencies (equivalently, the averaged categorical probabilities) to obtain the most likely token at that location.
\section{Experiments} \label{sec: Experiments}

\begin{wraptable}[13]{r}{0.48\textwidth}
\vspace{-1.\baselineskip}
\centering
\caption{Sudoku solving accuracy (\%). All methods except CTTT use the same backbone architecture.}
\label{tab:sudoku}
\setlength{\tabcolsep}{5pt}
\renewcommand{\arraystretch}{1.05}
\begin{tabular}{@{}lcc@{}}
\toprule
\textbf{Method} &
\textbf{\shortstack{Sudoku\\Extreme}} &
\textbf{\shortstack{Kaggle\\Unfiltered}} \\
\midrule
DFM (Euler)              & 31.6  & 44.5  \\
DFM + Top-K              & 68.8  & 89.2  \\
DFM + Top-K Margin       & 68.5  & 88.5  \\
DFM + ReMDM              & 51.3  & 71.9  \\
CTTT                      & --    & 98.9    \\
\midrule
Ours                     & \textbf{95.2}  & \textbf{99.9}  \\
\bottomrule
\end{tabular}
\vspace{-0.8em}
\end{wraptable}

All methods share the same Transformer architecture ($8$ blocks, $8$ attention heads, rotary position embeddings) trained with AdamW and a polynomial masking schedule. Architecture dimensions, time-conditioning, and training hyperparameters differ between tasks and are detailed in~\cref{app:setup}.

\subsection{Sudoku}\label{subsec:sudoku}
Sudoku is a standard logic puzzle played on a $9\times 9$ grid, partitioned into nine $3\times 3$ subgrids. 
Some cells are pre-filled as \emph{clues}, and the goal is to fill the remaining cells with digits $1$--$9$ 
such that each row, each column, and each $3\times 3$ subgrid contains every digit exactly once.

\begin{wrapfigure}[11]{r}{0.3\textwidth}
  \vspace{-1.\baselineskip}
  \centering
  \includegraphics[width=\linewidth]{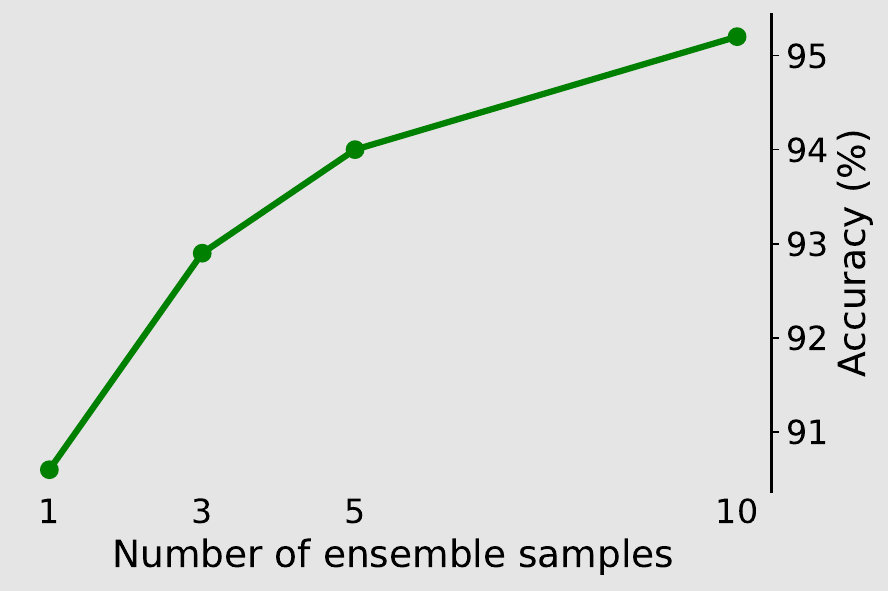}
  \caption{Sudoku accuracy vs.\ ensemble size.}
  \label{fig:ensemble_acc}
  \vspace{-0.5\baselineskip}
\end{wrapfigure}
\paragraph{Setup.}
We evaluate on two $9\!\times\!9$ Sudoku datasets of varying difficulty: \textbf{Sudoku-Extreme}~\cite{wang2025hierarchicalreasoningmodel}, a collection of $3{,}104{,}157$ challenging puzzles that typically require substantially more search/backtracking (i.e., ``guesses'') for classical solvers, and \textbf{Kaggle Unfiltered}~\cite{radcliffe2020_3m_sudoku}, a large-scale dataset of mixed difficulty with over one million puzzles (see~\cref{app:datasets} for details). All models are trained on Sudoku-Extreme: each input $x_0$ consists of the given clues and mask tokens for all remaining cells. We train our method for $3.8$ million steps, while the baseline is trained for $10$ million steps since its performance continued to improve over the longer schedule. We evaluate on $1{,}000$ held-out Sudoku-Extreme puzzles and additionally report generalization on the unseen Kaggle Unfiltered dataset.

\paragraph{Baselines.}

We compare the following methods, all using the same pretrained backbone unless noted otherwise. The baseline model is trained on the masking path \eqref{eq:masking path} and the cross entropy loss on $p_{1|t}$, see \eqref{eq:cross-entropy}. 
\emph{DFM}~\cite{gat2024discrete}: standard discrete flow matching with Euler sampling. For inference methods we are using 
\emph{Top-K} and \emph{Top-K Margin}~\cite{kim2025trainworstplanbest}: adaptive unmasking order based on model confidence; see~\cref{app:topk}.
\emph{ReMDM}~\cite{wang2026remaskingdiscretediffusionmodels}: inference-time remasking allowing previously decoded tokens to be reconsidered; see~\cref{app:remdm}.
\emph{GIDD}~\cite{vonruette2025generalizedinterpolatingdiscretediffusion}: pretraining with a generalized noise process mixing mask, uniform, and data tokens; see~\cref{app:gidd}. CTTT~\cite{giannoulis2026teachingtransformerssolvecombinatorial} is trained on its own randomly generated Sudoku dataset and evaluated on {Kaggle Unfiltered}. We report their published result.
\begin{wraptable}[9]{r}{0.3\textwidth}
  \vspace{1\baselineskip}
  \centering
  \caption{Average sampling steps vs.\ ensemble size.}
  \label{tab:ensemble_steps_only}
  \vspace{-0.25\baselineskip}
  \begin{tabular}{rc}
    \toprule
    \textbf{\# Ensemble} & \textbf{Avg.\ steps} \\
    \midrule
    1  & 64.2  \\
    3  & 105.1 \\
    5  & 123.5 \\
    10 & 146   \\
    \bottomrule
  \end{tabular}
  \vspace{-0.75\baselineskip}
\end{wraptable}

\paragraph{Ours.}
\cref{tab:sudoku} reports accuracy (the fraction of fully solved puzzles). We use 10 ensemble samples for inference (see \cref{subsec:Inference Strategies}), which yields a validity of $95.2\%$, and \cref{tab:ensemble_steps_only} reports the number of denoising steps. The {Kaggle Unfiltered} dataset is substantially easier then Sudoku Extreme: we solve all but one puzzle correctly. In \cref{fig:ensemble_acc}, we compare accuracy as a function of the ensemble size. Accuracy improves slightly with more ensemble samples, while the average number of steps increases sublinearly, from $62$ to $146$.


\subsection{Countdown}
\begin{wraptable}{r}{0.48\textwidth} 
\vspace{-0.8em} 
\centering
\caption{Countdown-4 (CD4). Baselines reproduced from Ye et al.~\cite{ye2025autoregressiondiscretediffusioncomplex}.}
\label{tab:countdown_cd4}
\setlength{\tabcolsep}{6pt}
\begin{tabular}{lcc}
\toprule
Method & Params & CD4 (\%) \\
\midrule
\multicolumn{3}{l}{\textit{Autoregressive}} \\
GPT-2 Scratch & 85M  & 45.8 \\
GPT-2 Scratch & 303M & 41.3 \\
Stream-of-Search & 250M & 54.2 \\
LLaMA & 7B  & 41.1 \\
LLaMA & 13B & 51.1 \\
\midrule
\multicolumn{3}{l}{\textit{Diffusion}} \\
VDM  & 85M & 73.4 \\
D3PM & 85M & 83.1 \\
RDM  & 85M & 87.0 \\
MGDM & 85M  & 91.5 \\
\midrule
\textbf{Ours} & 34M & \textbf{95.9} \\
\textbf{Ours} ($7\cdot 10^6$ steps) & 34M & \textbf{98.9} \\
\textbf{Small} & 11M & \textbf{92.1} \\
DFM & 34M & 87.5 \\
\bottomrule
\end{tabular}
\vspace{2pt}
{\footnotesize \textit{Note:} Non-ours numbers are taken from Ye et al.~\cite{ye2025autoregressiondiscretediffusioncomplex} (Table~1).}
\vspace{-0.8em} 
\end{wraptable}
Countdown~\cite{wikipedia_countdown_game_show_2024} is a numerical reasoning puzzle in which one is given a multiset of integers and a target value, and must produce a valid arithmetic expression using $+,-,\times,/$ (with exact integer division) that evaluates to the target while using each number at most once.
We focus on the \emph{Countdown-4} setting (four input numbers) and design our task and evaluation \emph{in the spirit of} ~\cite{ye2025autoregressiondiscretediffusioncomplex}, while making several implementation choices and constraints that differ from their exact setup. An example would be the numbers $86,28,13,31$ and the target $96$ which have to be combined as follows, $86+28=114$, $31-13=18$ and $114-18=96.$

\paragraph{Setup.}
We train on 500k examples generated following \cite{ye2025autoregressiondiscretediffusioncomplex, gandhi2024streamsearchsoslearning}, with target numbers ranging from 10 to 99.
In contrast to \cite{ye2025autoregressiondiscretediffusioncomplex}, we employ a base-$10{,}000$ tokenizer in which each integer from $0$ to $9{,}999$ is represented as a single token, rather than encoding numbers digit by digit.
All comparisons are taken from \cite{ye2025autoregressiondiscretediffusioncomplex}, which reports results for the diffusion models VDM \cite{kingma2023variationaldiffusionmodels}, D3PM \cite{austin2023structureddenoisingdiffusionmodels}, RDM \cite{zheng2024reparameterizeddiscretediffusionmodel}, as well as their method MGDM \cite{ye2025autoregressiondiscretediffusioncomplex}. The DFM baseline is implemented by us, and we train both models for $500\mathrm{k}$ steps.

\paragraph{Baseline.}
Our baseline performs comparably to the baseline reported in \cite{ye2025autoregressiondiscretediffusioncomplex} for MGDM.
We use 100 Euler steps in the discrete flow-matching solver \cite{gat2024discrete}.
 
\paragraph{Our Method.}
Using an ensemble of 5 models, our method achieves $95.9\%$ accuracy, substantially outperforming the DFM baseline ($87.5\%$) and exceeding the best reported non-ours validity ($91.5\%$). We also train a smaller variant with $11$M parameters for $500$k steps, which reaches $92.1\%$, still above MGDM despite using a much smaller network. With 5-model ensembling, the average number of steps is $24.7$. Using a single model, we obtain $93.8\%$ accuracy with only $7.2$ average steps. In the long-run we train for 7 millon steps, and reach an accuracy of $98.9\%$.

\section{Conclusion}
We propose a lightweight adaptation of discrete diffusion models for reasoning tasks. Our approach learns a Markov transition kernel on model-generated states, enabling self-correction by allowing tokens to be unmasked, kept, or re-masked during inference. Two additional heads provide the required control: a confidence head that guides corrective transitions and a learned stopping head that allocates a problem-dependent inference budget. Experiments on Sudoku and Countdown demonstrate that this adaptive, self-correcting inference substantially improves validity while often requiring only a small number of steps.

\paragraph{Acknowledgments.}
GS acknowledge funding by the German Research Foundation (DFG)
within the Excellence Cluster MATH+ and JC by project STE 571/17- 2 within the The Mathematics
of Deep Learning. GK acknowledges funding by the BMBF VIScreenPRO (ID: 100715327). CW
gratefully acknowledge funding by the DFG within the SFB “Tomography Across the
Scales”(STE571/19-1, projectnumber: 495365311)

\bibliography{iclr2026_delta}
\bibliographystyle{iclr2026_delta}

\newpage
\appendix

\section{Experimental Setup}\label{app:setup}

\subsection{Architecture}\label{app:architecture}

All models use a Transformer encoder with rotary position embeddings~(RoPE). The sequence of length~$d$ is embedded via a learned token embedding of dimension~$D_{\text{hidden}}$ plus a separate clue embedding indicating given positions.

\paragraph{Baseline model.}
Time conditioning follows the DiT design~\cite{peebles2023scalablediffusionmodelstransformers}: a sinusoidal timestep embedding is projected to $D_{\text{hidden}}$ via a two-layer MLP and injected through Adaptive Layer Normalization (AdaLN) in each Transformer block. Each block applies multi-head self-attention with RoPE, followed by a feed-forward network (expansion ratio~$4$, GELU activation), with residual connections and dropout. The final layer uses AdaLN followed by a linear projection to vocabulary logits.

\paragraph{Adaptive model.}
The backbone is identical except that AdaLN time conditioning is replaced by a simple additive time embedding. In addition, the model maintains two parallel hidden streams processed alongside the main token representations:
\begin{itemize}
    \item \textbf{Time prediction head.} An attention-pooled representation of the sequence is passed through a two-layer MLP with sigmoid output, predicting the scalar stopping time~$\tau_\theta \in [0,1]$ (\cref{eq:real time}).
    \item \textbf{Mixing prediction head.} A per-position representation, aggregated via masked attention pooling (excluding clue positions), is passed through an MLP producing the per-token confidence~$c_\theta^i \in [0,1]$.
\end{itemize}

\paragraph{Sudoku hyperparameters.}
Hidden dimension~$D_{\text{hidden}} = 512$, $8$ Transformer blocks, $8$ attention heads, dropout~$0.05$, vocabulary size~$K=10$ (digits $0$--$9$), sequence length~$d=81$.

\paragraph{Countdown hyperparameters.}
Hidden dimension~$D_{\text{hidden}} = 512$ and $D_{\text{hidden}} = 256$ for the small network in \cref{tab:countdown_cd4}, $8$ Transformer blocks, $8$ attention heads, dropout~$0.05$.
The vocabulary contains $10{,}007$ tokens: integers $0$--$9{,}999$ (one token each, base-$10{,}000$ encoding), operators $+$, $-$, $\times$, $/$, $=$, and a comma separator (six dedicated tokens), one EOS/padding token, and one mask token.
Sequence length~$d=40$.
Time conditioning follows the same baseline/adaptive split as Sudoku (AdaLN for the baseline, additive embedding for the adaptive model); the two auxiliary prediction heads are identical in design, with $\text{hidden\_dim}=256$ and $3$ MLP layers each.

\subsection{Training}\label{app:training}

\paragraph{Baseline training.}
We train with standard cross-entropy loss on $p_1^\theta$, applied at all positions. The masking schedule is polynomial $\kappa_t = t^2$. We use Adam with learning rate $3\!\times\!10^{-4}$, weight decay $10^{-4}$, linear warmup over $1000$ steps, gradient clipping at $1.0$, and batch size $256$. Logits are clipped to $[-50, 50]$.

\paragraph{Adaptive training.}
The loss combines three terms: (i)~standard cross-entropy on the denoiser $p_1^\theta$ (applied to all positions, no time weighting), (ii)~mixing loss encouraging confident commits only when correct (with $\varepsilon=0.05$, exponent~$k=1$), and (iii)~time prediction MSE. Training uses on-policy states: at each step, we sample $x_t$ from the masking path, take a single no-gradient forward step through the model to obtain $y \sim p_c^{\text{sg}(\theta)}(\cdot \mid x_t)$, and compute the loss on~$y$. Other hyperparameters match the baseline.

\subsection{Countdown Training}\label{app:countdown-training}

\paragraph{Baseline training.}
We minimise the standard cross-entropy loss on $p_1^\theta$, applied at all positions with no time reweighting.
The masking schedule is polynomial $\kappa_t = t^2$ (exponent~$2$).
Training uses AdamW with learning rate $5\!\times\!10^{-4}$, weight decay $0.02$, linear warmup over $1000$ steps, gradient clipping at $1.0$, batch size $128$, and logit clipping to $[-50, 50]$.
Both models are trained for $500\mathrm{k}$ steps.
At inference we use $100$ Euler steps.

\paragraph{Adaptive training.}
The loss follows~\eqref{eq:loss} with the same three terms as described for Sudoku.
The masking schedule is linear $\kappa_t = t$ (exponent~$1$).
On-policy states are generated via a short full-trajectory rollout of $5$ steps (applied with probability $0.1$): we sample $x_t$ from the masking path, unroll the model for $5$ steps to obtain a state~$z$, and compute the loss on~$z$ against the ground-truth solution.
All other hyperparameters (optimizer, learning rate, batch size, gradient clipping) match the Countdown baseline above.

\subsection{Datasets}\label{app:datasets}

\paragraph{Sudoku-Extreme.}
A curated collection of difficult $9\!\times\!9$ Sudoku puzzles from \cite{wang2025hierarchicalreasoningmodel}, each with $17$--$25$ given clues. The dataset contains approximately $50{,}000$ puzzles. Each puzzle is flattened to a sequence of length~$81$ with vocabulary $\{0,\dots,9\}$ where $0$ denotes a blank cell. Given clues are enforced throughout training and inference via a binary clue mask.

\paragraph{Kaggle Unfiltered.}
A large-scale dataset from \cite{radcliffe2020_3m_sudoku} of over one million $9\!\times\!9$ Sudoku puzzles with mixed difficulty, sourced from Kaggle. The same preprocessing and representation are used as for Sudoku-Extreme.

\paragraph{Countdown-4.}
We generate $500\mathrm{k}$ training examples following the setup
of~\cite{ye2025autoregressiondiscretediffusioncomplex, gandhi2024streamsearchsoslearning}:
four integers are drawn uniformly, and a target in $[10,99]$ is produced from a valid
arithmetic expression using $+,-,\times,/$ (exact integer division only).
Integers from $0$ to $9{,}999$ are each encoded as a single token
(base-$10{,}000$ tokeniser); operators ($+$, $-$, $\times$, $/$, $=$, comma) and the
EOS/padding symbol each receive a dedicated token, for a total vocabulary of
$10{,}007$ tokens (excluding the mask token, which is appended during training).
Each expression is serialised left-to-right and padded to a fixed sequence length of $40$.
The given numbers (clues) are provided as part of the input and are kept fixed
throughout training and inference.

\section{Comparison Methods}\label{app:comparisons}

\subsection{Top-K and Top-K Margin}\label{app:topk}

Top-K and Top-K Margin~\cite{kim2025trainworstplanbest} are inference-time strategies that modify the order in which masked positions are decoded, without retraining the model.

At each denoising step, a certainty score is computed for every masked position~$i$:
\begin{itemize}
    \item \textbf{Top-K:} $\text{certainty}_i = \max_{j \in [K]} p_\theta(x^i = j \mid x_t)$, i.e., the maximum predicted probability.
    \item \textbf{Top-K Margin:} $\text{certainty}_i = p_\theta(x^i = j_1 \mid x_t) - p_\theta(x^i = j_2 \mid x_t)$, where $j_1, j_2$ are the two most probable tokens. This more reliably captures uncertainty when multiple tokens have similar high probabilities.
\end{itemize}
The number of positions to unmask is $K_t = M_t \cdot (\alpha_s - \alpha_t) / (1 - \alpha_t)$, where $M_t$ is the current number of masked positions and $\alpha_t$ is the masking schedule. The top-$K_t$ positions by certainty are unmasked, with tokens sampled from the model's predicted distribution. Following~\cite{kim2025trainworstplanbest}, we add no Gumbel noise to the certainty scores and use multinomial sampling for token selection.

\subsection{ReMDM}\label{app:remdm}
Remasking Discrete Diffusion Models (ReMDM) modify the reverse posterior of masked diffusion to allow previously decoded tokens to be remasked, enabling iterative correction. For an unmasked token at position $i$, the reverse posterior is
\[
q_{\sigma}\!\left(z_s^{(i)} \mid z_t^{(i)} = x^{(i)}, x^{(i)}\right)
= (1-\sigma_t)\,\delta_{x^{(i)}} + \sigma_t\,\delta_m .
\]
A key result (Theorem 3.1) is that the marginal $q_{\sigma}(z_t \mid x)$ is identical to that of standard masked diffusion. This implies that ReMDM can be used as an inference-time sampler on top of a pretrained masked diffusion model without retraining. We use the rescale schedule
\[
\sigma_t = \eta\,\sigma_t^{\max},
\qquad
\sigma_t^{\max} = \min\!\left(1,\frac{1-\alpha_s}{\alpha_t}\right),
\]
with $\eta = 0.9$ as a chosen hyperparameter.

\subsection{GIDD Pretraining}\label{app:gidd}

Generalized Interpolating Discrete Diffusion~(GIDD)~\cite{vonruette2025generalizedinterpolatingdiscretediffusion} generalizes the masked diffusion noise process to include uniform noise alongside masking. The conditional forward process is:
\begin{align*}
    p_t^{\text{GIDD}}(\cdot \mid x_0, x_1) = \kappa_t^1\,\delta_{x_0}(\cdot) + \kappa_t^2\,p_u(\cdot) + \kappa_t^3\,\delta_{x_1}(\cdot),
\end{align*}
where $p_u$ is the uniform distribution over the vocabulary, $\sum_i \kappa_t^i = 1$, and the boundary conditions ensure interpolation from $x_0$ at $t\!=\!0$ to $x_1$ at $t\!=\!1$. The uniform component $\kappa_t^2$ allows training states to contain random incorrect tokens (rather than only mask or correct tokens), exposing the model to error states during training. However, these are uniformly random errors, not the model's own systematic mistakes.

We train a GIDD baseline using the same backbone architecture, with the hybrid noise schedule recommended in~\cite{vonruette2025generalizedinterpolatingdiscretediffusion}.

\section{Discrete Flow Matching}\label{sec:DFM}
For convenience, we reconsider discrete flow matching with our notation based on \cite{gat2024discrete,lipman2024flowmatchingguidecode}. Just for the notation, for a random variable $X_t$, $t \in [0,1]$, having a law with density $p_{X_t}$, we just write $p_t$, and accordingly
\begin{align*}
p_t (x) &= p_{X_t}(x), \quad 
p_{t|0,1}(x|x_0,x_1) = p_{X_t|(X_0,X_1) = (x_0,x_1)}(x), \\
p_{0,1|t}(x_0,x_1|x) &= p_{(X_0,X_1)|X_t =x} (x_0,x_1), \quad 
p_{1|t}(x|x_t) = p_{X_1|X_t = x_t} (x)
\end{align*}

For a function $u:[0,1] \times  \cS \times \cS \to \mathbb R$ which is continuous in time on $(0,1)$ and any initial probability mass function $p_{\text{init}}: \cS \to [0,1]$, the \textbf{linear ODE} system (Kolmogorov equation)
\begin{equation} \label{eq:kolmogorov}
\dot p_t(x)  = \sum_{y \in \cS} u_t(x,y) p_t(y),  x \in \cS \qquad p_0 = p_{\text{init}}
\end{equation}
has a unique solution. Supposing the \textbf{rate condition}
\begin{equation} \label{eq:rate}
\sum_{x \in \cS} u_t(x,y) = 0 \quad \text{for all } y \in \cS, 
\qquad u_t(x,y) \ge 0 \quad \text{for } x \not = y,
\end{equation}
we obtain that $p_t$ remains a PMF. In this case, we have
$$
\sum_{y \in \cS} u_t(x,y) p_t(y) 
= \sum_{y \not = x } \underbrace{u_t(x,y) p_t(y)}_{j_t(x,y)} - \sum_{y \not = x }\underbrace{u_t(y,x) p_t(x)}_{j_t(y,x)} 
 =  \sum_{y \in \cS } [j_t(x,y) - j_t(y,x)]
$$
with flux $j_t(x,y) = u_t(x,y) p_t(y)$ from $y$ to $x$, so that
$(p_t,u_t)$ fulfill the \textbf{continuity equation}
\begin{equation} \label{eq:CE}
\dot p_t + \text{div} (p_t u_t) = 0, \qquad p_0 = p_{\text{init}}
\end{equation}
where 
$$\text{div} (j_t)(y) = - \sum_{y \in \cS} \big( j_t(x,y) - j_t(y,x) \big).$$ 
In the following, we are exclusively interested in velocity fields fulfilling the rate condition \eqref{eq:rate}.

We want to find a flow $p_t$ from the initial distribution to the target distribution, $p_1 = p_{\text{tar}}$.
The aim of FM is to learn the velocity field $u_t$ and then to use e.g. the Euler scheme of the ODE \eqref{eq:kolmogorov}  to sample from $p_1$.

Let $\pi$ be a coupling between the source distribution and the target distribution.
Then we have by the law of total probability that
\begin{equation} \label{eq:prob_full}
p_t(x) = \sum_{x_0,x_1 \in \cS} p_{t|0,1}(x|x_0,x_1) \pi(x_0,x_1).
\end{equation}
Using 
$$p_{t|0,1}(y|x_0 ,x_1) \, \pi(x_0,x_1) = p_{0,1|t}(x_0, x_1|y) \, p_t(y),$$
the velocity field  to $p_t$ can be deduced as
\begin{align*} 
\dot p_t(x) 
&=  \sum_{x_0,x_1 \in \cS} \dot p_{t|0,1}(x|x_0 x_1) \pi(x_0,x_1)
\\ 
&= \sum_{x_0,x_1 \in \cS} \sum_{y \in \cS} u_t(x,y|x_0,x_1) p_{t|0,1} (y|x_0 x_1) \; \pi(x_0,x_1)
\\
&=  \sum_{y \in \cS} \sum_{x_0,x_1 \in \cS}  u_t(x,y|x_0,x_1) p_{0,1|t}(x_0 x_1|y)\,  p_t(y).
\end{align*}
Thus, the velocity field belonging to $p_t$ is 
\begin{equation} \label{eq:velo_full}
u_t(x,y) = \sum_{x_0,x_1 \in \cS}  u_t(x,y|x_0,x_1) p_{0,1|t}(x_0 x_1|y) .
\end{equation}
To manage the computational burden, we are only interested in conditional PMF that respect the special structure of our state space $\cS$ and restrict ourselves to independent components, i.e., with PMFs $p_t^i: \cT \to [0,1]$,  we consider
$$
p_{t|0,1}(x|x_0,x_1) = \prod_{i=1}^d p_{t|0,1}^i(x^i|x_0,x_1)
$$
The PMFs $p_t^i(\cdot|x_0, x_1)$ can be defined in several ways. Here we choose
\begin{equation} \label{eq:path}
p_{t|0,1}^i(x^i|x_0,x_1) = (1-\kappa_t) \delta (x^i,x_0^i) + \kappa_t \delta (x^i,x_1^i).
\end{equation}
with a monotone increasing, continuously differentiable  schedule function $\kappa_t:[0,1] \to [0,1]$ fulfilling $\kappa_0 = 0$ and $\kappa_1 = 1$.
The velocity fields corresponding to  $p_t^i(\cdot|x_0,x_1)$  can be simply computed by
\begin{align*}
 \frac{\d}{\d t}  p_{t|0,1}^i(x^i|x_0,x_1) 
 &= \dot \kappa_t (  \delta (x^i,x_1^i) - \delta (x^i,x_0^i) )\\
 &= \dot \kappa_t (  \delta (x^i,x_1^i) - \frac{p_{t|0,1}^i(x^i|x_0,x_1) + \kappa_t \delta(x^i,x_1^i)}{1- \kappa_t}\\
 &= \frac{\dot \kappa_t}{1- \kappa_t} \big(\delta_{x_1^i}(x^i) - p_{t|0,1}^i(x^i|x_0,x_1) \big)\\
 &=  \frac{\dot \kappa_t}{1- \kappa_t} \sum_{y^i \in \cT} \big(\delta(x^i,x_1^i) - \delta(x^i,y^i) \big) p_{t|0,1}^i(y^i|x_0,x_1). 
\end{align*}
Thus, {\color{blue} a } velocity field belonging to $p_{t|0,1}^i(\cdot|x_0,x_1)$ in \eqref{eq:kolmogorov} is given by
\begin{equation} \label{eq:down}
u_t^i(x^i,y^i|x_0,x_1) = \frac{\dot \kappa_t}{1- \kappa_t}\big(\delta(x^i,x_1^i) - \delta(x^i,y^i) \big).
\end{equation}
Then the vector field $u_t(x,y|x_0,x_1)$ belonging to $p_t(\cdot|x_0,x_1)$ is 
\begin{align} \label{eq:cond_flow}
u_t(x,y|x_0,x_1) &= u_t^1(x^1,y^1|x_0,x_1) \, \boldsymbol{\iota_1}
\; + \ldots + \; u_t^d(x^d,y^d|x_0,x_1) \, \boldsymbol{\iota_d}\\
&= \sum_{i=1}^d  u_t^i(x^i,y^i|x_0,x_1)  \, \boldsymbol{\iota_i}
\end{align}
where
$$
 \boldsymbol{\iota_i} = \prod_{j=1 \atop j \not = i}^d \delta ( x^j,y^j). 
$$
Here $\boldsymbol{\iota_i}$ can be seen as a pointer that $u_t^i(x^i,y^i|x_0,x_1)$ appears everywhere in $u_t(x,y|x_0,x_1)$ where $x^j=y^j$ for all $j\not = i$.
For a general derivation of vector fields of product measures 
fulfilling a continuity equation see \cite{duong2026telegraphersgenerativemodelkac}.
\\
Reason: for fixed $x_0,x_1$, let 
$q_t = p_{t|0,1}(\cdot|x_0,x_1)$ and $v_t = u_t(\cdot|x_0,x_1)$.
By the product rule we obtain for $q_t(x) = q^1_t(x^1) \ldots q_t^d(x^d)$ that
\begin{align*}
\dot q_t(x) 
&= 
\dot q_1(x^1) \frac{q_t(x)}{q_t^1(x^1)} + \ldots +  \dot q_d(x^d) \frac{q_t(x)}{q_t^d(x^d)}\\
&= 
\sum_{y^1 \in \cT} v_t^1(x^1,y^1) q_t^1(y^1) \frac{q_t(x)}{q_t^1(x^1)}
+ \ldots +  \sum_{y^d \in \cT} v_t^d(x^d,y^d) q_t^d(y^d) \frac{q_t(x)}{q_t^d(x^d)}\\
&= 
\sum_{y^1 \in \cT} \ldots \sum_{y^d \in \cT} 
 \Big( v_t^1(x^1,y^1) \prod_{j\not = 1} \delta (x^j,y^j)
\; + \ldots + \; v_t^d(x^d,y^d) \prod_{j\not = d} \delta ( x^j,y^j) \Big) q_t(y)\\
&= 
 \sum_{y \in \cS} v_t(x,y) q_t(y).
\end{align*}
By \eqref{eq:velo_full}, \eqref{eq:cond_flow} and \eqref{eq:down}, we obtain
\begin{align*}
u_t(x,y) 
&= \sum_{x_0,x_1 \in \cS}   \Big(\sum_{i=1}^d u_t^i(x^i,y^i|x_0,x_1) \, \boldsymbol{\iota_i} \Big)
p_{0,1|t}(x_0,x_1|{\color{red} y})\\
&= \frac{\dot \kappa_t}{1- \kappa_t} \sum_{i=1}^d \sum_{x_0,x_1 \in \cS}
\big(\delta(x^i,x_1^i) - \delta(x^i,y^i) \big)  p_{0,1|t}(x_0,x_1|y)  \, \boldsymbol{\iota_i} .
\end{align*}
Since
\begin{align*}
\sum_{x_0,x_1 \in \cS}
\big(\delta(x^i,x_1^i) - \delta(x^i,y^i) \big)  p_{0,1|t}(x_0,x_1|y) 
&= \sum_{x_1 \in \cS }\delta(x^i,x_1^i)  p_{1|t}(x_1|y) - \delta(x^i,y^i)
\\
&= \sum_{x_1^i\in \cT} p_{1|t}^i(x_1^i|y) \delta(x^i,x_1^i) - \delta(x^i,y^i), 
\end{align*}
it follows finally that
\begin{align}\label{eq:vel_final}
 u_t(x,y) &=  
 \sum_{i=1}^d  \frac{\dot \kappa_t}{1- \kappa_t}\big( \sum_{x_1^i \in \cT}p_{1|t}^i(x_1^i| y) \delta (x^i,x_1^i) - \delta(x^i,y^i) \big) \, \boldsymbol{\iota_i} \\
 &=
 \sum_{i=1}^d \underbrace{ \frac{\dot \kappa_t}{1- \kappa_t} \Big(p_{1|t}^i(x^i|y) - \delta(x^i,y^i) \Big)}_{u_t^i(x^i,y)} \, \boldsymbol{\iota_i}.
\end{align}
Therefore we have $u_t$ if we know $p^i_{1|t}$. 
The Kullback-Leibler divergence of two PMFs $p,q:\cT \to [0,1]$ is defined by 
$$\kl(p,q) = \sum_{x \in \cT} \log p(x) p(x) - \log q(x) p(x)$$
and their \textbf{cross entropy} by
\begin{align}\label{eq:cross-entropy}
{\rm CE}(p,q) = - \sum_{x \in \cT} \log q(x) p(x) = \kl (p,q) - \sum_{x \in \cT} \log p(x) p(x),
\end{align}
where we suppose that
$p(x) = 0$ whenever $q(x) = 0$, $x \in \cT$.

To learn for $i=1,\ldots,d$ the PMF $p_{1|t}^i(\cdot| y) \delta (\cdot,x_1^i)$,
we minimize the expectation value of the cross entropy 
\begin{align*}
\text{CE} \left(p_{1|t}^i(\cdot|x_t) ,  p_{1|t}^{\theta,i}(\cdot|x_t )  \right) 
&= - \sum_{z \in \cT} p^i_{1|t} (z|x_t) \log p_{1|t}^{i,\theta}(z|x_t)
\\
&= 
- \sum_{z \in \cT} \sum_{x_0,x_1} 
\underbrace{p^i_{1|t,0,1}(z|x_t,x_0,x_1)}_{\delta(z,x_1^i)} \pi(x_0,x_1)
\log p_{1|t}^{i,\theta}(z|x_t)\\
&= -  \sum_{x_0,x_1} \log p_{1|t}^{i,\theta}(x_1^i|x_t)
\end{align*}
i.e., we maximize for every $i =1,\ldots,d$ the loss function
\begin{align*}
\cL(\theta) 
&= \mathbb E_{t \sim \cU(0,1), (x_0,x_1) \sim \pi, x_t \sim p_{t|0,1}}  
\Big[ \sum_{i=1} ^d
\log \big(p_{1|t}^{\theta,i}(x_1^i|x_t ) \big) \Big].
\end{align*}
For inference, the \textbf{Euler forward scheme of the ODE}
\begin{equation} \label{eq:euler}
p_{t+h} (x) = p_t (x) + h \sum_{y\in \cS} u_t^\theta(x,y) p_t(y)
\end{equation} 
is computed with the learned velocity $u_t^\theta(x,y)$. The velocity field $u_t$ satisfying the rate condition defines a CTMC on $S$. For sampling, this CTMC is discretized via the Euler scheme $$X^i_{t+h} \sim \delta_{X^i_t}(\cdot) + h\, u^i_t(\cdot, X_t).$$

\end{document}